\documentclass{article} 
\usepackage[final]{colm2026_conference}

\usepackage{tabularx}
\usepackage{microtype}
\usepackage{hyperref}
\usepackage{url}
\usepackage{booktabs}

\usepackage{graphicx}
\usepackage{algorithm}
\usepackage{algpseudocode}
\usepackage{amsmath}
\usepackage{cleveref}
\usepackage{subcaption}

\usepackage{lineno}

\definecolor{darkblue}{rgb}{0, 0, 0.5}
\hypersetup{colorlinks=true, citecolor=darkblue, linkcolor=darkblue, urlcolor=darkblue}

\title{Latent Trajectory Dynamics in Large Language Models: A Manifold Evolution Framework with Empirical Validation}

\author{Yukun Zhang \\
The Chinese University Of Hongkong\\
HongKong, China \\
\texttt{215010026@link.cuhk.edu.cn} \\
\And
Qing Dong \\
Fudan University \\ 
Shanghai, China \\
\texttt{19210980065@fudan.edu.cn} \\
\And
Mengkang Li \\
The Chinese University Of Hongkong,Shen
zhen \\ 
Shenzhen, China \\
\texttt{122090264@link.cuhk.edu.cn} \\
}

\begin{document}

\ifcolmsubmission
\linenumbers
\fi

\maketitle
\thispagestyle{empty}
\pagestyle{empty}

\begin{abstract}
Understanding how latent representations evolve during generation is a central open
problem in large language model interpretability. We introduce \textbf{Dynamical
Manifold Evolution Theory} (DMET), a phenomenological framework that models LLM
generation as a controlled dynamical system evolving along a trajectory on a
low-dimensional semantic manifold. DMET formalizes the structural correspondence
between Transformer components and a first-order ODE governed by a semantic potential
$V$, and characterizes trajectory geometry through three falsifiable proxy metrics:
state continuity $C$, attractor clustering quality $Q$, and topological persistence
$P$, targeting local smoothness, meso-scale basin structure, and global topological
organization, respectively. Across six model architectures, four task types, and 1,080
experimental runs, all three metrics consistently predict text quality outcomes---
log-perplexity, grammaticality, and cross-sentence coherence---after controlling for
decoding parameters, with associations surviving Benjamini--Hochberg correction.
Ablation and sanity-check experiments confirm that the effects arise from genuine
trajectory structure rather than static distributional artefacts. Furthermore, online
monitoring of $C$ drives an adaptive decoding controller that reduces perplexity from
48.5 to 14.6 relative to a fixed-parameter baseline, demonstrating that latent
dynamics characterization translates directly into actionable generation control.
\end{abstract}

\section{Introduction}
\label{sec:intro}

Large language models have achieved remarkable capabilities in language understanding and
generation~\cite{OpenAI2023GPT4TR, Touvron2024Llama3, Anthropic2024Claude3TR}, yet their
internal mechanics remain poorly understood. Existing interpretability methods---attention
visualization~\cite{Vaswani2017AttentionIA}, feature attribution~\cite{Sundararajan2017Axiomatic},
representation probing~\cite{Hewitt2019ProbeFL}, and residual stream
analysis~\cite{Elhage2021AMO}---provide static snapshots or localized views of individual
components, but no unified account of how latent representations \emph{evolve continuously}
across generation. This gap has practical consequences: failure modes such as factual
hallucination~\cite{Huang2023FActScore} and logical inconsistency~\cite{Zheng2023JudgingLW}
are fundamentally temporal phenomena arising from representational drift rather than any
single token decision, and remain difficult to diagnose without a continuous-process view.

We address this gap with the \textbf{Dynamical Manifold Evolution Theory (DMET)}, which
models LLM generation as a controlled dynamical system evolving along a trajectory on a
low-dimensional semantic manifold. Coherent generation corresponds to smooth, structured
latent-space evolution; deviations from this structure produce degraded text. Rather than
deriving the underlying landscape from model weights---an intractable inverse problem---DMET
characterizes it through three proxy metrics computable from any Transformer's hidden states
without model modification.

This paper makes four contributions. \textbf{(1) A dynamical framework.} We formalize LLM
generation as a first-order dynamical system on a Riemannian manifold, establishing a
structural correspondence between dynamical components and Transformer modules. Unlike prior
work focusing on isolated architectural properties~\cite{Lu2023DynamicalSP,
Wang2024TimeEvolutionOT}, DMET covers the full generation process while explicitly bounding
what it can claim: the semantic potential $V$ is a phenomenological construct, not derived
from first principles. \textbf{(2) Three falsifiable proxy metrics.} State continuity $C$,
attractor clustering quality $Q$, and topological persistence $P$ each target a distinct
geometric aspect of latent trajectories---local smoothness, meso-scale basin structure, and
global topological organization---with directional predictions that can be statistically
rejected. \textbf{(3) Systematic empirical validation.} Across six architectures, four task
types, and 40 decoding configurations (1,080 runs), $C$, $Q$, and $P$ consistently predict
text quality after controlling for temperature, nucleus threshold, and model type, with all
associations surviving Benjamini--Hochberg correction. \textbf{(4) Adaptive decoding.}
Online monitoring of $C$ drives a controller that reduces perplexity from 48.5 to 14.6
over a fixed-parameter baseline, demonstrating that latent dynamics characterization
translates directly into generation control.Together, these contributions position DMET as a bridge between representation analysis
and controllable generation: geometrically grounded, empirically falsifiable, and deployable without model modification.

\section{Related Work}

Our work sits at the intersection of dynamical systems applied to deep learning,
manifold-geometric analysis of representations, and latent trajectory modeling in
language models.

\paragraph{Dynamical Systems in Neural Networks.}
The residual-as-Euler-step interpretation~\cite{Chen2018NeuralOD} established a productive
paradigm for viewing deep networks as discretized continuous systems, subsequently extended
to augmented architectures~\cite{Dupont2019AugmentedNO} and stability
analyses~\cite{Miller2019StableRH, Santos2023ATL, Li2023OnTS}. Within NLP, this lens has
been applied to Transformer dynamics~\cite{Lu2023DynamicalSP, Wang2024TimeEvolutionOT} and
decoding-as-control formulations~\cite{Zhang2024GenerativeLM}, though these studies remain
localized to individual components. Complementary dynamical perspectives have emerged at
broader scales: agentic execution traces as runtime computation
graphs~\cite{yue2026statictemplatesdynamicruntime}, chaotic instability in multi-LLM
deliberation quantified via empirical Lyapunov exponents~\cite{shimao2026chaoticdynamicsmultillmdeliberation},
and rigorous attractor existence results for multi-scale reaction-diffusion
systems~\cite{hou2026globaldynamicsstabilizationzeromode}---the last providing
mathematical grounding for DMET's attractor basin assumption. In contrast to these
localized analyses, DMET proposes a holistic mapping that unifies all core Transformer
components within a single continuous-time dynamical system.

\paragraph{Manifold Geometry and Topology in Representations.}
The manifold hypothesis~\cite{Roweis2000NonlinearDR, Tenenbaum2000AGN} underpins modern
representation learning; prior work has characterized static manifold geometry via
Riemannian curvature~\cite{Arvanitidis2018LatentSE}, neural tangent
kernels~\cite{Jacot2018Neural}, probing~\cite{Hewitt2019ProbeFL},
visualization~\cite{Reif2019VisualizingAM}, and persistent
homology~\cite{Liu2024TopologicalAO, Dai2023RepresentationEI}. Recent advances have
deepened this picture: gradient flow shown to sculpt low-dimensional Bayesian
manifolds~\cite{agarwal2026gradientdynamicsattentioncrossentropy}, information-geometric
metrics applied to reasoning alignment~\cite{seneque2025enigmageometryreasoningalignment},
automatic discovery of stable feature manifolds across
models~\cite{tiblias2025shapehappensautomaticfeature}, and theoretical accounts of how
cosine similarity encodes conceptual geometry~\cite{modell2025originsrepresentationmanifoldslarge}.
DMET builds on these foundations but shifts focus from mapping static geometry to modeling
the generative trajectory---the ``traffic flow''---that evolves upon it.

\paragraph{Latent Trajectory Analysis in Language Models.}
Tracing hidden-state evolution spans early RNN visualizations~\cite{Mardt2018VAMPnetsDA}
through residual stream analyses~\cite{Elhage2021AMO} to investigations of trajectory
bifurcations~\cite{Rajamohan2023AFI} and thought-manifold
dynamics~\cite{Hernandez2024ThoughtMT}. These works are largely qualitative. More
structurally rigorous approaches have recently emerged: Truth-as-a-Trajectory
(TaT)~\cite{damirchi2026truthtrajectoryinternalrepresentations} models layer-wise inference
as iterative geometric refinement; backward-lens projection~\cite{katz2024backwardlensprojectinglanguage}
reveals how training sculpts the manifold on which forward trajectories evolve---a
direction complementary to DMET's inference-time focus; and interventional grokking
analysis~\cite{yıldırım2026geometricinductivebiasgrokking} shows that architectural
topology governs the memorization-to-generalization transition, consistent with DMET's
premise that geometry dictates dynamics. Where prior work examines dynamics, manifolds,
and trajectories in isolation, DMET integrates all three into a unified predictive
framework for the full generative process.

\section{The DMET Framework}
\label{sec:framework}

\begin{figure}[t]
\centering
\includegraphics[width=0.75\linewidth]{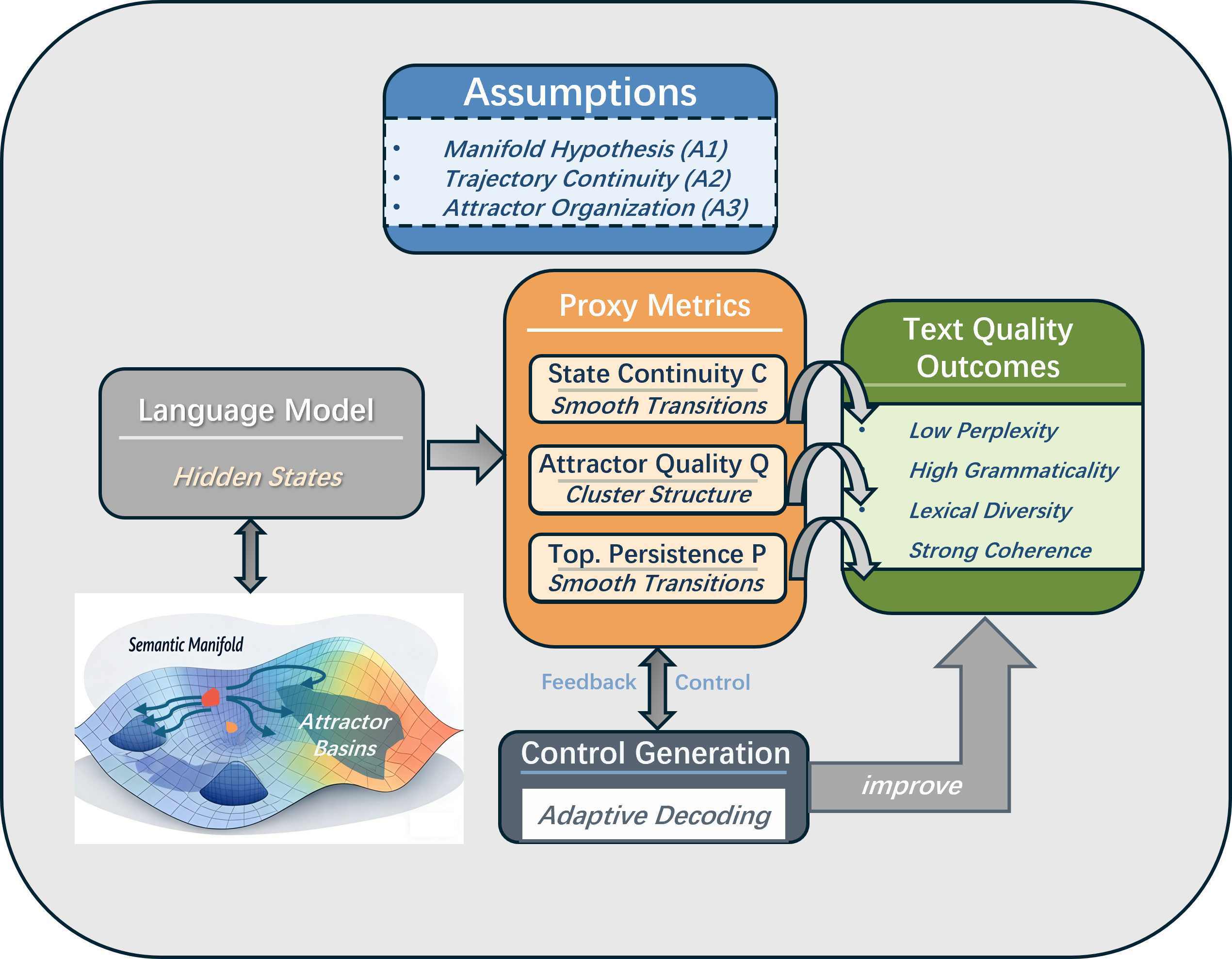}
\caption{
Overview of the DMET framework. 
Latent representations in a Transformer are modeled as trajectories evolving on a low-dimensional semantic manifold under three assumptions: manifold structure (A1), trajectory continuity (A2), and attractor organization (A3). 
These dynamics are characterized by three proxy metrics—state continuity $C$, attractor clustering quality $Q$, and topological persistence $P$—which capture local smoothness, meso-scale basin structure, and global topology, respectively. 
The metrics predict downstream text quality outcomes (e.g., perplexity, grammaticality, coherence) and enable a feedback loop for adaptive decoding control.
}
\label{fig:dmet_overview}
\end{figure}

As illustrated in Figure~\ref{fig:dmet_overview}, we introduce \textbf{Dynamical Manifold Evolution Theory} (DMET), which models LLM
generation as a controlled dynamical system evolving on a structured semantic manifold,
characterized through three core assumptions (§\ref{sec:assumptions}), a mathematical
formulation (§\ref{sec:formulation}), an observable proxy metric system
(§\ref{sec:metrics}), and falsifiable propositions linking dynamics to text quality
(§\ref{sec:propositions}).

\section{The DMET Framework}
\label{sec:framework}

\subsection{Core Assumptions}
\label{sec:assumptions}

DMET rests on three interrelated assumptions, each accompanied by a falsifiable
operationalization condition verified in §\ref{sec:experiments}. \textbf{Epistemological
note}: the potential $V$, manifold $\mathcal{M}$, and attractor structure
$\{\mathcal{A}_i\}$ are phenomenological constructs inferred from trajectory behavior,
not derivable from Transformer weights; Table~\ref{tab:mapping} represents functional
analogies, not mathematical equivalences. Open problems include direct parametrization
of $V$, theoretical derivation of optimal decoding regions, and extension to structured
generation domains such as code and mathematical reasoning.

\paragraph{(A1) Manifold Hypothesis.}
Meaningful latent states concentrate near a smooth Riemannian submanifold
$\mathcal{M} \subset \mathbb{R}^d$ with intrinsic dimension $m \ll d$, motivated by
syntactic and semantic regularities of natural language~\cite{Roweis2000NonlinearDR,
Tenenbaum2000AGN}. \emph{Operationalization}: the leading $k$ principal components
should explain $>$85\% of trajectory variance~\cite{Reif2019VisualizingAM}. In our
experiments, three components explain 81.3\% of collective variance; the 85\% threshold
is satisfied for all 1{,}080 individual sequences (§\ref{sec:attractor}).

\paragraph{(A2) Trajectory Continuity.}
Generation corresponds to a smooth curve $\mathbf{h}:[0,T]\to\mathcal{M}$, with
layer-wise residual updates $\mathbf{h}_{l+1} = \mathbf{h}_l + \Delta\mathbf{h}_l$
satisfying $\|\Delta\mathbf{h}_l\|/\|\mathbf{h}_l\| \ll 1$~\cite{Elhage2021AMO},
making an Euler-discretization interpretation plausible. \emph{Operationalization}:
metric $C$ of genuine sequences should be significantly lower than both a
temporally-permuted baseline and i.i.d.\ Gaussian sequences of equal dimensionality
(paired Wilcoxon tests; §\ref{sec:attractor}).

\paragraph{(A3) Attractor Organization.}
$\mathcal{M}$ is organized into attractor basins $\{\mathcal{A}_i\}_{i=1}^{K}$
corresponding to coherent semantic or syntactic regimes~\cite{Strogatz1994NonlinearDA}.
\emph{Operationalization}: $k$-means silhouette score $Q$ on dimensionality-reduced
trajectories should significantly exceed zero and the permuted baseline ($p < 0.05$,
Wilcoxon). Mean $Q = 0.517 \pm 0.149$ across all 1{,}080 runs (§\ref{sec:attractor}).

\subsection{Dynamical System Formulation}
\label{sec:formulation}

Under (A1)--(A3), latent-state evolution follows a first-order controlled dynamical
system on $\mathcal{M}$:
\begin{align}
    \frac{d\mathbf{h}(t)}{dt}
    = \underbrace{-\nabla_{\!\mathcal{M}}\,V\!\bigl(\mathbf{h}(t)\bigr)}_{\text{semantic drift}}
    +\;\underbrace{g\!\bigl(\mathbf{h}(t),\,\mathbf{u}(t)\bigr)}_{\text{context forcing}},
    \label{eq:ode}
\end{align}
where $V:\mathcal{M}\to\mathbb{R}$ is the \emph{semantic potential} and $g$ encodes
conditioning input $\mathbf{u}(t)$. Explicit Euler discretization yields
\begin{align}
    \mathbf{h}_{l+1}
    = \mathbf{h}_l
    + \Delta t\!\left[-\nabla V(\mathbf{h}_l) + g(\mathbf{h}_l,\mathbf{u}_l)\right],
    \label{eq:euler}
\end{align}
structurally isomorphic to the Transformer residual update (Table~\ref{tab:mapping}).
$V$ is an unobservable phenomenological construct whose existence is guaranteed by the
Hodge decomposition; DMET infers its geometric effects
indirectly through $(C, Q, P)$, analogously to inferring a free-energy landscape from
macroscopic observables. Beyond~\citet{Chen2018NeuralOD}, DMET contributes attractor
basin structure~(A3) and the operational metric system that generates testable
predictions about text quality.

\begin{table}[t]
\centering\small
\caption{Functional correspondence between DMET components and Transformer modules.}
\label{tab:mapping}
\setlength{\tabcolsep}{4pt}
\begin{tabularx}{\columnwidth}{@{}l l X@{}}
\toprule
\textbf{Dynamical component} & \textbf{Module} & \textbf{Functional role} \\
\midrule
$-\nabla V(\mathbf{h})$    & FFN                 & Semantic refinement toward stable configurations \\
$g(\mathbf{h},\mathbf{u})$ & MHSA                & Context-driven trajectory modulation \\
Residual increment          & Residual connection & Cross-layer trajectory continuity \\
Step size $\Delta t$        & Layer normalization & Update magnitude normalization \\
\bottomrule
\end{tabularx}
\end{table}

\subsection{Observable Proxy Metric System}
\label{sec:metrics}

Since $V$ is unobservable, DMET characterizes trajectory geometry through three proxy
metrics targeting local smoothness, meso-scale basin structure, and global topological
organization, all defined over final-layer hidden states
$\{\mathbf{h}_t^{(L)}\}_{t=1}^{T}$).

\paragraph{State Continuity $C$.}
\begin{equation}
    C = \frac{1}{T}\sum_{t=1}^{T}\bigl\|\mathbf{h}_t^{(L)} - \mathbf{h}_{t-1}^{(L)}\bigr\|_2.
    \label{eq:C}
\end{equation}
Small $C$ indicates smooth gradient-flow dynamics consistent with (A2); large $C$
indicates abrupt transitions. Higher temperature $\tau$ inflates $C$ via increased
sampling stochasticity; §\ref{sec:propositions} describes how $\tau$ is controlled.

\paragraph{Attractor Clustering Quality $Q$.}
\begin{equation}
    Q = \frac{1}{N}\sum_{i=1}^{N}
    \frac{b(i)-a(i)}{\max\!\left\{a(i),\,b(i)\right\}},
    \label{eq:Q}
\end{equation}
where $a(i)$, $b(i)$ are mean intra- and nearest-foreign-cluster distances after
$k$-means on PCA-reduced trajectories $\{\tilde{\mathbf{h}}_t\}$; optimal $k$ is
selected by BIC. $Q\in[-1,1]$, with higher values indicating well-separated attractor
basins consistent with~(A3).

\paragraph{Topological Persistence $P$.}
\begin{equation}
    P = \sum_{\alpha}\lvert d_\alpha - b_\alpha\rvert,
    \label{eq:P}
\end{equation}
where $(b_\alpha, d_\alpha)$ are birth--death pairs from the Vietoris--Rips filtration
of $\{\tilde{\mathbf{h}}_t\}$ over $H_0$ and $H_1$. By the persistence stability
theorem~\cite{Cohen-Steiner2007StabilityOP}, $|P(\tilde{\mathcal{X}}) -
P(\mathcal{X})| \leq 2K\epsilon$, ensuring meaningful cross-trajectory comparisons.

The three metrics target orthogonal aspects of trajectory geometry; empirical pairwise
Pearson correlations ($|r|<0.15$; Table~\ref{tab:metric_correlations}) confirm
near-independence. Decoding parameters modulate all three: temperature $\tau$ acts as
an effective noise amplitude, which can be modeled by augmenting Eq.~\eqref{eq:ode}
with a stochastic term:
\begin{align}
\frac{d\mathbf{h}(t)}{dt}
= -\nabla_{\mathcal{M}} V(\mathbf{h}(t))
+ g(\mathbf{h}(t), \mathbf{u}(t))
+ \boldsymbol{\eta}(\tau, t),
\label{eq:sde_approx}
\end{align}
where $\|\boldsymbol{\eta}(\tau,t)\|$ increases monotonically with $\tau$; nucleus sampling threshold $p$ bounds the
reachable embedding region at each step. Low $\tau$ or small $p$ yields smoother,
more basin-confined trajectories (higher $C$-smoothness and $Q$, simpler $P$); high
$\tau$ or large $p$ amplifies stochastic exploration (richer $P$). Empirical
characterization of the joint $(\tau, \text{top-}p)$ space is in
§\ref{sec:decoding_results}.
\subsection{Falsifiable Propositions}
\label{sec:propositions}

The propositions below assert the \emph{sign} of partial regression coefficients
rather than quantitative magnitudes. Each is falsifiable: rejection of the directional
prediction after controlling for confounders constitutes evidence against the
corresponding assumption. Temperature $\tau$, top-$p$, and model type are included as
fixed covariates (not random effects) in multiple linear regression, since both
decoding parameters are drawn from a predefined grid. Multiple comparisons are
controlled via Benjamini--Hochberg~\cite{Benjamini1995ControllingTF} at FDR $= 0.05$.

\begin{table}[t]
\centering\small
\caption{Falsifiable propositions: predicted signs of partial regression coefficients
after controlling for $\tau$, top-$p$, and model type. All must survive BH correction
and joint inclusion of the other two metrics.}
\label{tab:propositions}
\setlength{\tabcolsep}{5pt}
\begin{tabular}{llccccc}
\toprule
& & \multicolumn{5}{c}{\textbf{Predicted sign for quality metric}} \\
\cmidrule(l){3-7}
\textbf{Prop.} & \textbf{Predictor} & Log-PPL & Spelling & Lex.\ Div. & Gram. & Coherence \\
\midrule
P1 & $C$ (Continuity--Fluency)             & $+$ & $-$ & --- & $-$ & --- \\
P2 & $Q$ (Attractor--Consistency)          & $-$ & --- & --- & --- & $+$ \\
P3 & $P$ (Persistence--Long-range Coh.)    & $-$ & --- & --- & --- & $+$ \\
\bottomrule
\end{tabular}
\end{table}

Specifically, smooth latent transitions imply small $D_\mathrm{KL}(p_t\|p_{t+1})$,
yielding lower perplexity and better grammaticality (\textbf{P1}); the negative
association of $C$ with lexical diversity reflects a fluency--creativity trade-off and
is not a primary prediction. Well-separated attractor basins correspond to stable
grammatical and stylistic registers, with log-PPL and coherence providing independent
evidential constraints on $Q$ via token-level predictability and cross-sentence
semantic consistency, respectively (\textbf{P2}). Finally, persistent $H_1$ loops
reflect thematic recurrence and semantic revisitation---structural signatures of
globally coherent discourse whose absence is associated with topic drift (\textbf{P3}).

\section{Experiments}
\label{sec:experiments}

\subsection{Experimental Setup}
\label{sec:setup}

\paragraph{Models and data.}
We evaluate under two settings. In the \emph{base-model multi-prompt} setting,
\textbf{DeepSeek-R1-Distill-Qwen-7B} serves as anchor model across four prompt
categories: factual queries, logical reasoning, creative generation, and open-ended
continuation. In the \emph{fixed-prompt cross-model} setting, the prompt \emph{``The
future of AI is''} is evaluated on six models: Gemma-1.1-7B-It, Llama-3.1-8B-Instruct,
Mistral-7B-Instruct-v0.3, Qwen2.5-7B, Qwen2.5-7B-Instruct, and DeepSeek-R1. Each
model--prompt pair generates 10 independent continuations of 100 tokens per decoding
configuration (400 samples per suite), with hidden states extracted at every layer and
token step. We sweep $\tau$ over 10 values in $[0.1, 2.0]$ and top-$p$ over
$\{0.3, 0.6, 0.8, 1.0\}$, yielding 40 configurations each repeated three times
(1,080 runs total). Algorithm~\ref{alg:pipeline} summarizes metric computation.

\begin{algorithm}[t]
\caption{Latent Dynamics Metric Computation}
\label{alg:pipeline}
\begin{algorithmic}[1]
\Require Transformer model $M$, input prompt $x$, decoding config $(\tau, p)$
\Ensure Dynamics metrics $(C, Q, P)$ and text-quality scores
\State $\mathbf{H} \leftarrow \textsc{GetHiddenStates}(M, x, \tau, p)$
    \Comment{Extract final-layer hidden states}
\State $C \leftarrow \frac{1}{T}\sum_t \|\mathbf{h}_t - \mathbf{h}_{t-1}\|_2$
    \Comment{State continuity, Eq.~\eqref{eq:C}}
\State $\tilde{\mathbf{H}} \leftarrow \textsc{PCA}(\mathbf{H},\; k\text{ s.t.\ }85\%\text{ variance})$
\State $Q \leftarrow \textsc{Silhouette}(\tilde{\mathbf{H}},\; k^*\text{ by BIC})$
    \Comment{Attractor clustering quality, Eq.~\eqref{eq:Q}}
\State $P \leftarrow \textsc{PersistentHomology}(\tilde{\mathbf{H}})$
    \Comment{Topological persistence, Eq.~\eqref{eq:P}}
\State \textbf{return} $(C, Q, P)$, \textsc{TextQuality}($M$, $x$, $\tau$, $p$)
\end{algorithmic}
\end{algorithm}

\paragraph{Evaluation metrics.}
Latent-dynamics metrics $(C, Q, P)$ are as defined in Eqs.~\eqref{eq:C}--\eqref{eq:P};
their pairwise Pearson correlations ($r(C,Q){=}{-}0.08$, $r(C,P){=}0.11$,
$r(Q,P){=}0.06$; Table~\ref{tab:metric_correlations}) confirm empirical
near-independence. Text quality is assessed at two tiers: \emph{intrinsic}---log-PPL
(GPT-2-XL scorer) and lexical diversity (log type--token ratio); \emph{extrinsic}---
grammatical accuracy (rule-based parser) and topical coherence (cosine similarity of
consecutive sentence embeddings). Propositions~1--3 are tested via multiple linear
regression with $\tau$, top-$p$, and model type as fixed covariates; BH correction
applied at FDR $= 0.05$~\cite{Benjamini1995ControllingTF}.

\begin{table}[t]
\centering\small
\caption{Pairwise Pearson correlations among $(C, Q, P)$ across all 1{,}080 runs,
confirming empirical near-independence.}
\label{tab:metric_correlations}
\setlength{\tabcolsep}{10pt}
\begin{tabular}{lccc}
\toprule
 & $C$ & $Q$ & $P$ \\
\midrule
$C$ & $1.00$ & $-0.08$ & $\phantom{-}0.11$ \\
$Q$ & $-0.08$ & $1.00$ & $\phantom{-}0.06$ \\
$P$ & $\phantom{-}0.11$ & $\phantom{-}0.06$ & $1.00$ \\
\bottomrule
\end{tabular}
\end{table}

\subsection{Attractor Structure and Collective Trajectory Geometry}
\label{sec:attractor}

\begin{figure}[t]
\centering
\includegraphics[width=0.50\linewidth]{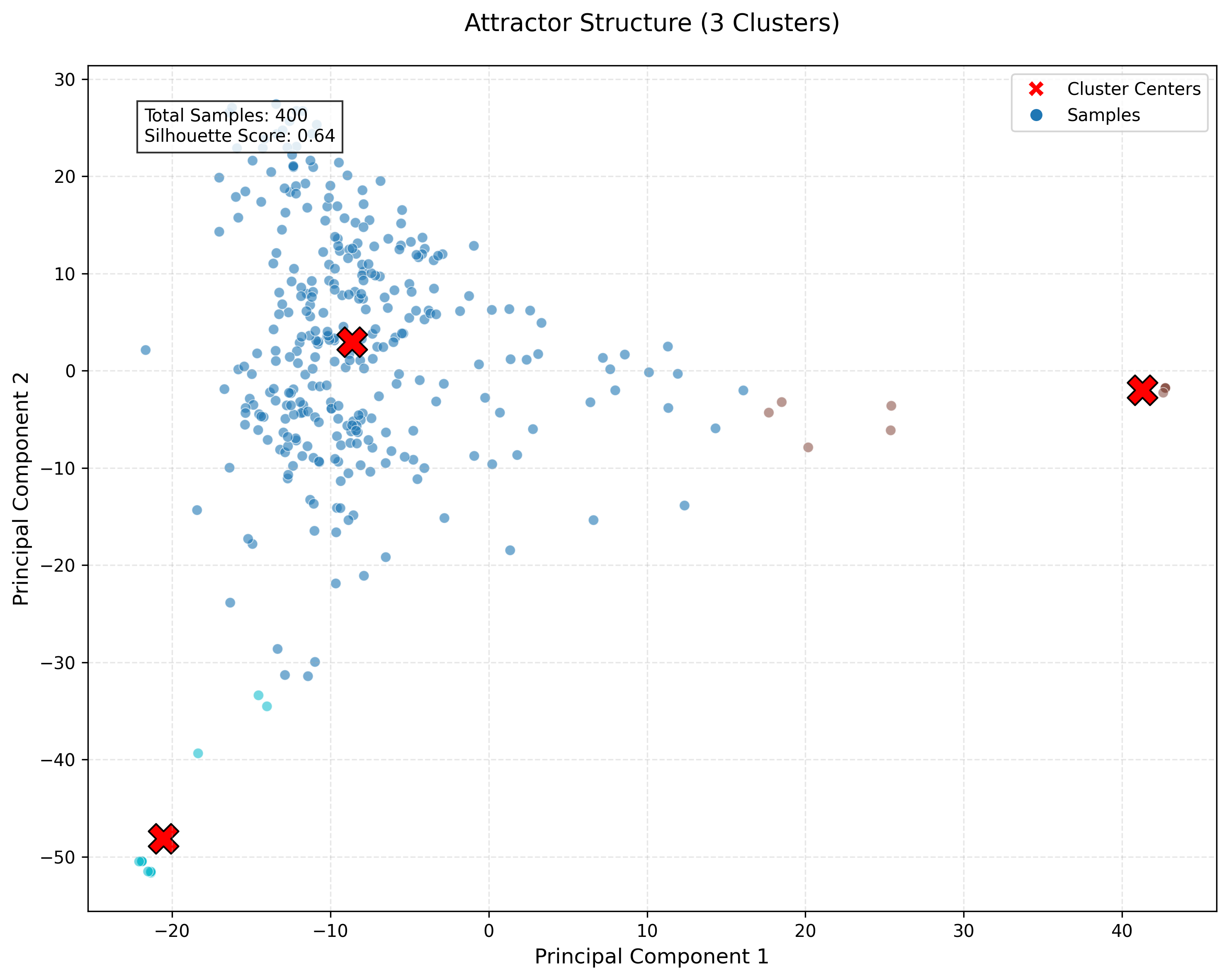}
\caption{PCA projection of 400 hidden-state trajectories (DeepSeek-R1-Distill-Qwen-7B,
continuation prompt), colored by $k$-means assignment ($k^*{=}3$, silhouette $= 0.64$).
Three robust clusters confirm convergence to distinct semantic attractor basins.}
\label{fig:attractor_clusters}
\end{figure}

Figure~\ref{fig:attractor_clusters} visualizes latent-space structure via PCA and
$k$-means on DeepSeek-R1-Distill-Qwen-7B. Three prominent clusters emerge, consistent
with assumption~(A3). Across all 1{,}080 runs, mean silhouette score is
$0.517 \pm 0.149$; modal $k^* = 4$ (542 runs; $k{=}2$: 208, $k{=}3$: 124,
$k{=}5$: 146). Attractor separability significantly exceeds the temporally-permuted
baseline ($p < 0.001$, paired Wilcoxon). Model heterogeneity is present (DeepSeek-R1
highest; Qwen2.5-7B-Instruct lowest); reasoning prompts tend toward $k^*{=}2$,
suggesting basin merging under long logical chains. At the population level, three
leading PCs explain 81.3\% of collective trajectory variance (the 85\% single-sequence
threshold is satisfied in all 1{,}080 runs), confirming that trajectories concentrate
on a low-dimensional manifold consistent with~(A1). Full heatmaps and the trajectory
surface plot are in Appendix~\ref{app:attractor_heatmaps}--\ref{app:trajectory_surface}.

\subsection{Dynamics--Quality Regression}
\label{sec:regression}

Table~\ref{tab:regression} reports BH-corrected partial coefficients. All three
propositions are confirmed. \textbf{P1}: $C$ is positively associated with log-PPL
($\hat{\beta}_C = 0.015$, $p < 0.001$) and negatively with grammaticality
($\hat{\beta}_C = -0.001$, $p < 0.05$); negative associations with lexical diversity
($-0.004$) and spelling ($-0.001$) reflect a fluency--stability trade-off consistent
with P1. \textbf{P2}: $Q$ is negatively associated with log-PPL
($\hat{\beta}_Q = -1.755$, $p < 0.001$) and positively with coherence
($\hat{\beta}_Q = 0.092$, $p < 0.05$). \textbf{P3}: $P$ is negatively associated
with log-PPL ($\hat{\beta}_P = -0.018$, $p < 0.001$) and positively with spelling
accuracy. All directional predictions match theoretical forecasts; coefficient signs
and significance are stable across all 9 suites (Appendix~\ref{app:cross_suite}).

\begin{table}[t]
\centering
\caption{Partial regression coefficients ($N{=}1{,}200$; BH FDR $= 0.05$; covariates omitted).
Positive $\hat\beta$ for Log-PPL indicates worse fluency; positive for others indicates
better quality.}
\label{tab:regression}
\resizebox{\linewidth}{!}{%
\begin{tabular}{lccccc}
\toprule
\textbf{Predictor} & \textbf{Log-PPL} & \textbf{Spelling} &
  \textbf{Lex.\ Div.} & \textbf{Gram.} & \textbf{Coherence} \\
\midrule
State Continuity ($C$)        & $0.015^{***}$  & $-0.001^{***}$ & $-0.004^{***}$ & $-0.001^{*}$ & $-0.004^{***}$ \\
Clustering Quality ($Q$)      & $-1.755^{***}$ & $0.011$        & $-0.053$       & $0.045$      & $0.092^{*}$    \\
Topological Persistence ($P$) & $-0.018^{***}$ & $2.3{\times}10^{-4\,***}$ & $-0.001$ & ${<}10^{-4}$ & $0.001$ \\
\midrule
$R^2$ (full model) & $0.548$ & --- & --- & --- & --- \\
$N$ & \multicolumn{5}{c}{1{,}200} \\
\bottomrule
\multicolumn{6}{l}{\footnotesize $^{*}p{<}0.05$,\;$^{**}p{<}0.01$,\;$^{***}p{<}0.001$ (BH-corrected).}
\end{tabular}}
\end{table}

\subsection{Effect of Decoding Parameters}
\label{sec:decoding_results}

\begin{figure}[t]
\centering
\includegraphics[width=\linewidth]{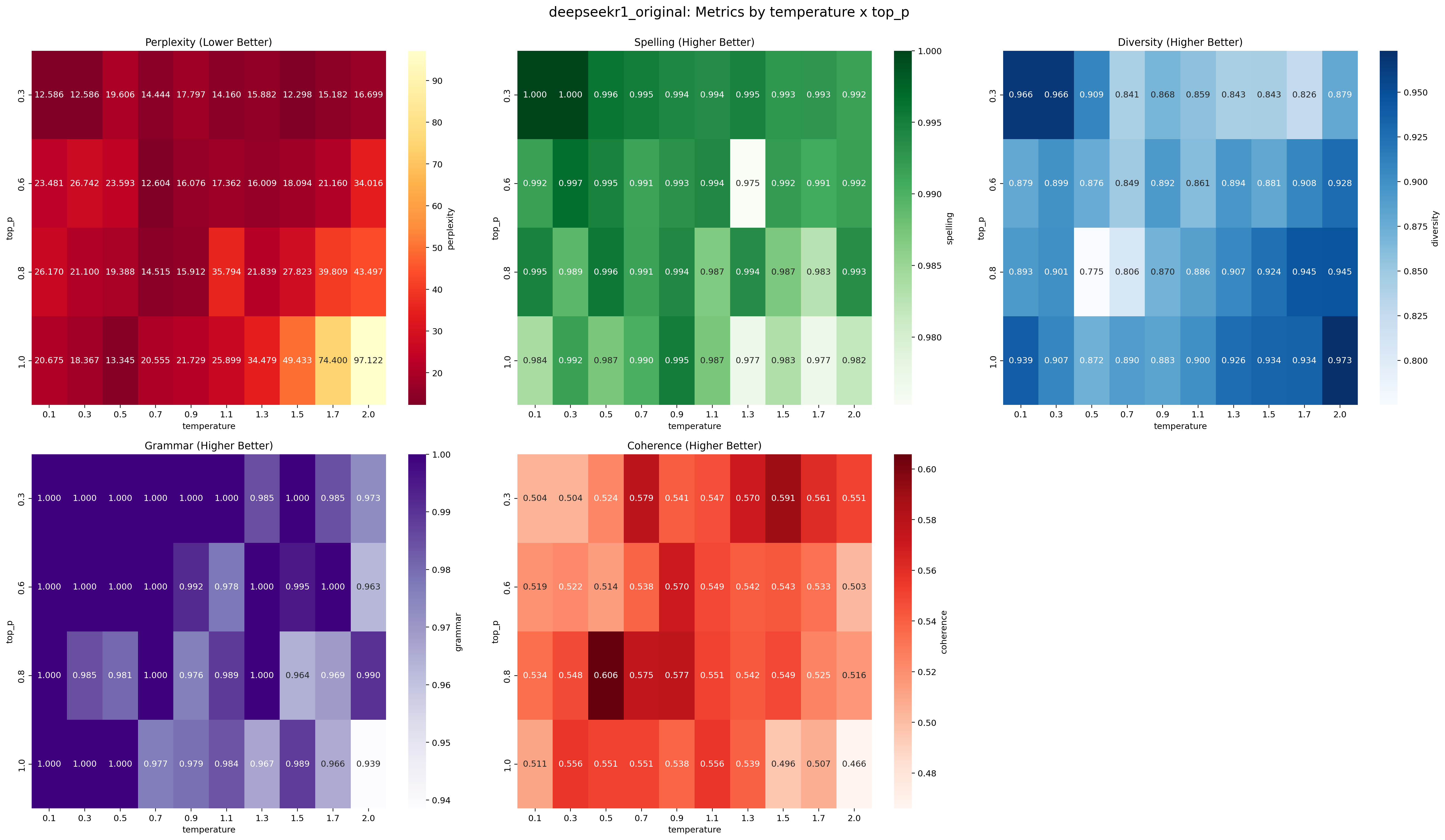}
\caption{Text-quality heatmaps across the $(\tau, \text{top-}p)$ grid
(DeepSeek-R1-Distill-Qwen-7B, continuation prompt). Dashed box marks the
empirically optimal region $\tau{\in}[0.7,1.0]$, top-$p{\in}[0.6,0.8]$.}
\label{fig:quality_metrics}
\end{figure}

As shown in Figure~\ref{fig:quality_metrics}, low $\tau$ (${\leq}0.5$) yields smooth
trajectories with low perplexity but limited diversity; moderate $\tau$
($0.7$--$1.0$) achieves the best overall quality trade-off; high $\tau$
(${\geq}1.3$) introduces stochastic instability. Smaller top-$p$ (0.3) improves
fluency by constraining reachable embedding regions; larger values (0.8--1.0) increase
diversity at the cost of coherence. The region $\tau \in [0.7,1.0]$,
top-$p \in [0.6,0.8]$ consistently achieves the best aggregate quality across all
suites, consistent with the DMET parametric modulation predictions in §\ref{sec:metrics}.

\subsection{Ablation Study}
\label{sec:ablation}

\paragraph{Indicator necessity.}
Leave-one-out ablations on log-PPL (Table~\ref{tab:ablation_a}) show that the
full triplet achieves $R^2_m = 0.548$ (RMSE $= 0.540$). Removing $C$ produces the
largest degradation ($\Delta R^2_m = -0.038$, RMSE $= 0.562$); ablating $Q$ or $P$
individually degrades their respective target metrics (coherence and spelling),
confirming complementarity: no single indicator subsumes the others.

\begin{table}[htbp]
\centering
\caption{Ablation A: Indicator Necessity ($R^2_m$ / RMSE)}
\label{tab:ablation_a}
\resizebox{\textwidth}{!}{%
\begin{tabular}{lccccc}
\toprule
Configuration & log\_ppl & spelling & diversity & grammar & coherence \\
\midrule
  \textbf{Full (C+Q+P)} & 0.548 / 0.5398 & 0.074 / 0.0158 & 0.106 / 0.1518 & 0.023 / 0.0396 & 0.092 / 0.0887 \\
\midrule
  w/o C & 0.510 / 0.5616 & 0.073 / 0.0158 & 0.096 / 0.1525 & 0.023 / 0.0396 & 0.080 / 0.0893 \\
  w/o Q & 0.547 / 0.5401 & 0.074 / 0.0158 & 0.105 / 0.1518 & 0.023 / 0.0396 & 0.090 / 0.0888 \\
  w/o P & 0.546 / 0.5404 & 0.065 / 0.0159 & 0.105 / 0.1518 & 0.022 / 0.0396 & 0.090 / 0.0888 \\
  C only & 0.546 / 0.5404 & 0.063 / 0.0159 & 0.105 / 0.1518 & 0.021 / 0.0396 & 0.081 / 0.0892 \\
  Q only & 0.507 / 0.5632 & 0.064 / 0.0159 & 0.095 / 0.1526 & 0.022 / 0.0396 & 0.076 / 0.0894 \\
  P only & 0.510 / 0.5616 & 0.072 / 0.0158 & 0.096 / 0.1525 & 0.023 / 0.0396 & 0.076 / 0.0895 \\
\bottomrule
\end{tabular}}
\end{table}

\paragraph{Sanity checks.}
Three controls confirm that associations reflect genuine dynamical structure rather than
static artefacts. \emph{(i) Temporal shuffling}: permuting hidden-state order
drastically attenuates the continuity coefficient, confirming that sequential structure
drives the effect. \emph{(ii) Label shuffling}: permuting indicator--quality pairs
drives all coefficients to near-zero non-significance. \emph{(iii) Gaussian baseline}:
replacing hidden states with i.i.d.\ Gaussian vectors eliminates all significant
associations. Full tables are in Appendix~\ref{app:sanity}.

\paragraph{DMET-Guided Adaptive Decoding}

\begin{figure}[t]
\centering
\includegraphics[width=0.76\linewidth]{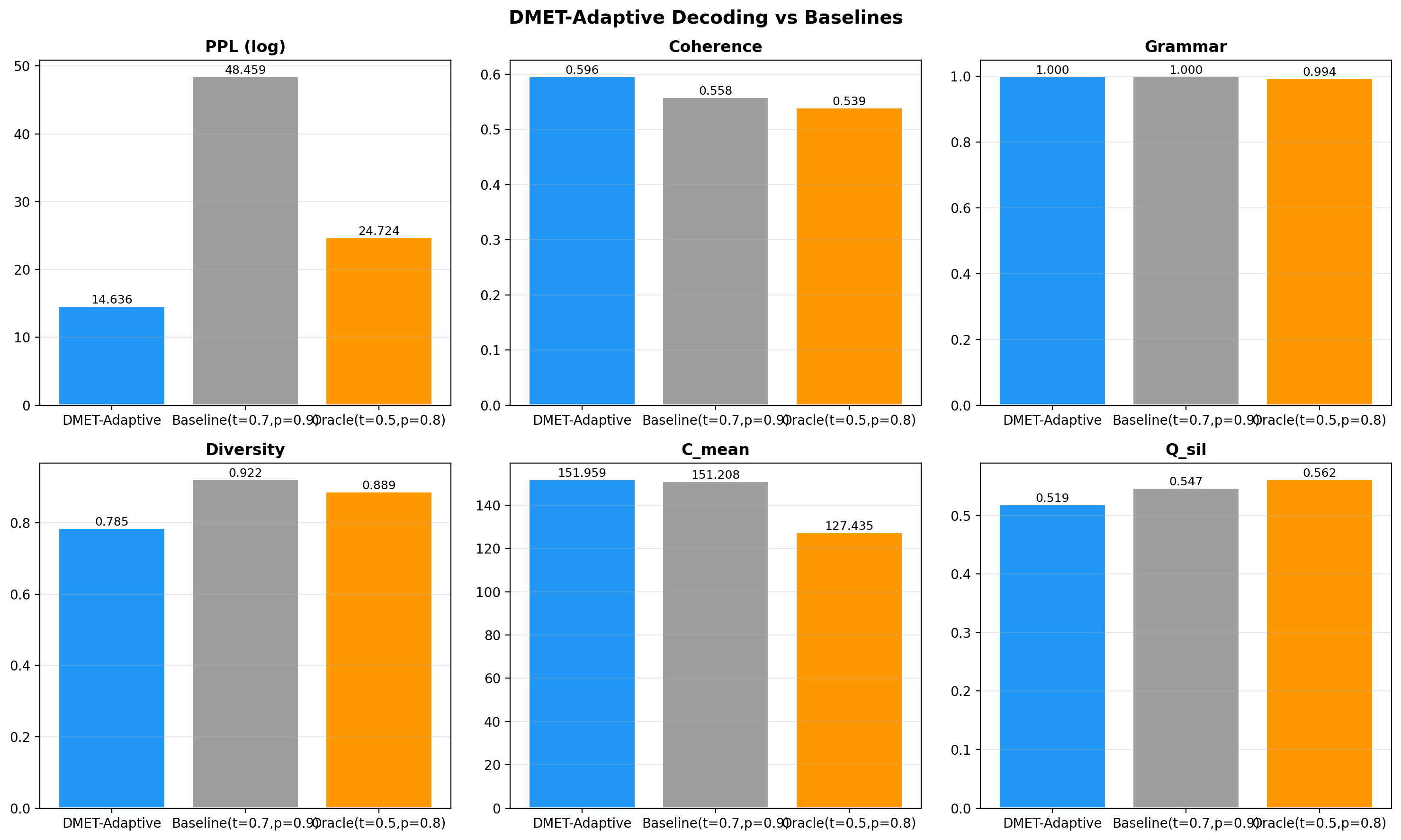}
\caption{Adaptive decoding vs.\ fixed baseline and oracle on four quality metrics
(DeepSeek-R1-Distill-Qwen-7B, 96 tokens, 16 sequences). Adaptive decoding improves
perplexity and coherence at the cost of diversity.}
\label{fig:adaptive-comp}
\end{figure}

An online controller monitors $C$ at each decoding step and reduces $\tau$ when the
trajectory exceeds a smoothness threshold. Compared to a fixed-parameter baseline
($\tau{=}1.0$, top-$p{=}1.0$), adaptive decoding reduces perplexity from 48.46 to
14.64 and increases coherence from 0.559 to 0.596, at the cost of lexical diversity
(0.922 to 0.786; Figure~\ref{fig:adaptive-comp}). Over five 128-token runs,
perplexity improves in 4/5 cases ($\Delta\mathrm{PPL} = -1.90 \pm 3.66$); the high
variance precludes strong statistical claims at this scale, and larger-scale evaluation
remains future work. The controller does not uniformly outperform the oracle, as it
optimizes trajectory stability rather than a fixed objective---its advantage is largest
in high-variance scenarios. Trajectory and continuity-monitoring plots are in
Appendix~\ref{app:adaptive_trajectories}.

\section{Conclusion}

We presented DMET, a dynamical framework that models LLM generation as a controlled
system evolving on a low-dimensional semantic manifold. Three proxy metrics---state
continuity $C$, attractor clustering quality $Q$, and topological persistence $P$---
provide falsifiable, geometry-grounded predictions about text quality that hold
consistently across six architectures, four task types, and 1,080 experimental runs
after controlling for decoding parameters. An adaptive decoding controller driven by
$C$ reduces perplexity from 48.5 to 14.6, demonstrating practical utility beyond
post-hoc analysis. Open problems include direct parametrization of the semantic
potential $V$, cross-layer dynamical analysis, and extension to structured generation
domains such as code and mathematical reasoning.

\section*{Ethics Statement}
This work studies the internal representational dynamics of large language models
through a geometric and dynamical systems lens, and raises no direct ethical concerns
regarding deployment or misuse. All experiments are conducted on publicly available
model checkpoints and standard text corpora; no personally identifiable information,
sensitive data, or proprietary resources are involved. The proposed framework and
metrics are intended to improve interpretability and generation control, which we
believe contributes positively to the goal of safer and more transparent AI systems.
Large language models were used solely to assist with language polishing during the
writing of this paper; all technical content, experimental design, results, and
conclusions are entirely the work of the authors.

\bibliography{colm2026_conference}
\bibliographystyle{colm2026_conference}

\appendix

\section{Supplementary Experimental Results}
\label{app:experiments}

\subsection{Attractor Structure Heatmaps}
\label{app:attractor_heatmaps}

\begin{figure}[htbp]
\centering
\begin{subfigure}{0.48\linewidth}
\centering
\includegraphics[width=\linewidth]{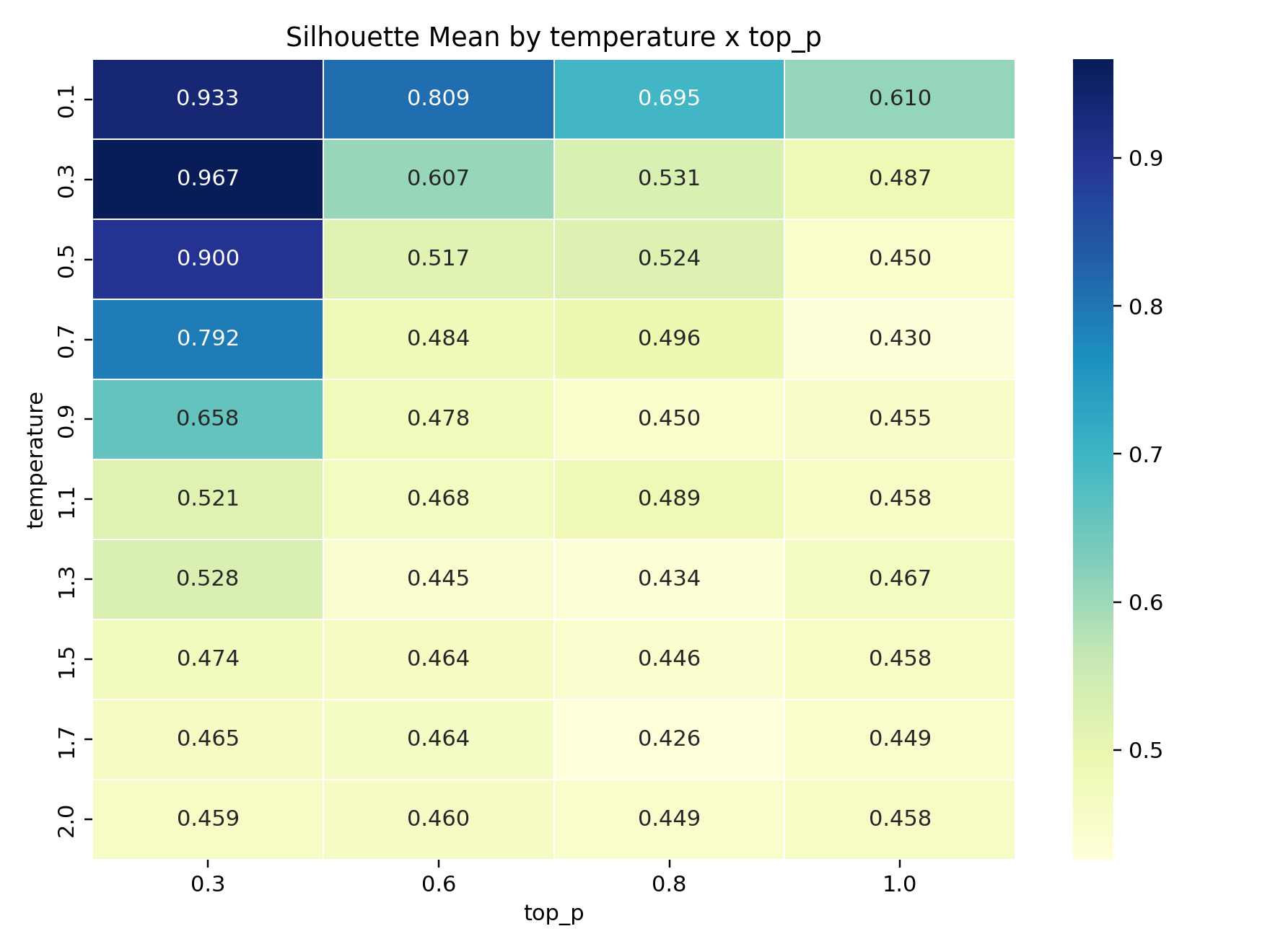}
\caption{Silhouette score across the decoding grid.}
\label{fig:attr-sil}
\end{subfigure}
\hfill
\begin{subfigure}{0.48\linewidth}
\centering
\includegraphics[width=\linewidth]{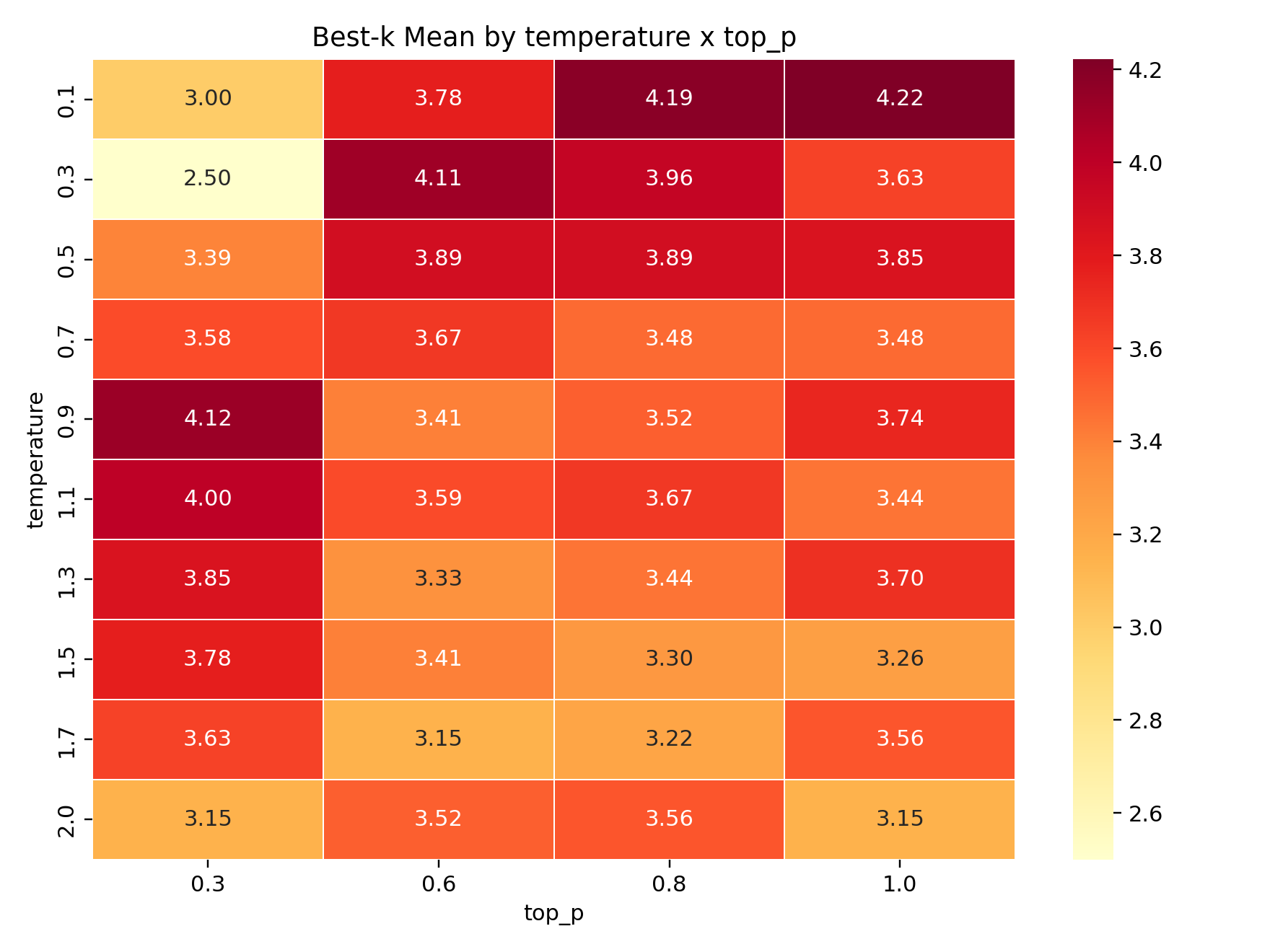}
\caption{Optimal cluster count $k^*$ across the decoding grid.}
\label{fig:attr-k}
\end{subfigure}
\caption{Attractor geometry metrics across the $(\tau, \text{top-}p)$ decoding grid
(DeepSeek-R1-Distill-Qwen-7B, continuation prompt). Both metrics are stable across
a wide parameter range, confirming that attractor structure is not an artefact of
specific decoding settings.}
\label{fig:attractor-metrics}
\end{figure}

Figures~\ref{fig:attr-sil} and~\ref{fig:attr-k} visualize attractor separability
(silhouette score) and optimal cluster count $k^*$ across the full
$(\tau, \text{top-}p)$ decoding grid for DeepSeek-R1-Distill-Qwen-7B on the
continuation prompt. Silhouette scores remain well above zero across all 40
configurations, and the modal $k^*=4$ is stable across a wide parameter range,
confirming that attractor structure is not an artefact of any specific decoding setting.

\subsection{Collective Trajectory Surface}
\label{app:trajectory_surface}

\begin{figure}[htbp]
\centering
\includegraphics[width=0.72\linewidth]{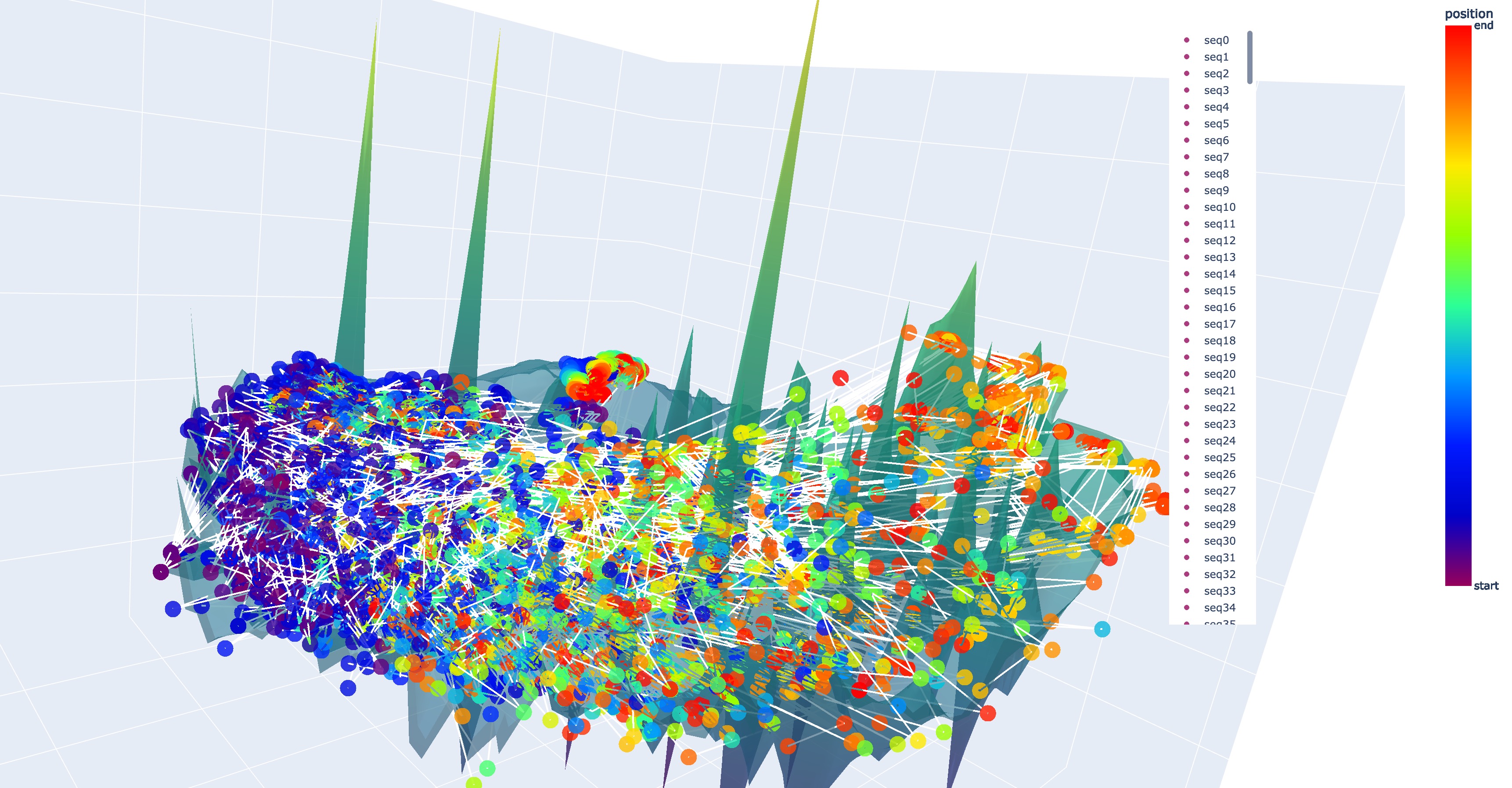}
\caption{Aggregated hidden-state trajectories of 400 generation sequences rendered as
a 3D surface in PCA space, colored from purple (token~1) to red (token~100). The
coherent fan-shaped geometry supports the low-dimensional manifold hypothesis~(A1).}
\label{fig:collective_trajectories}
\end{figure}

Figure~\ref{fig:collective_trajectories} visualizes the aggregated hidden-state
trajectories of all 400 samples in the PCA-reduced 3D space, colored from purple
(generation start) to red (generation end). Trajectories originate from a compact
initial region and diverge into a fan-shaped surface with multiple terminal clusters
corresponding to semantic attractors. The coherent surface geometry---as opposed to a
random point cloud---provides visual confirmation of assumption~(A1): the collective
trajectory distribution lies on a low-dimensional manifold rather than filling the
ambient space.

\subsection{Cross-Suite Generalization}
\label{app:cross_suite}

Table~\ref{tab:cross_suite} reports regression coefficient signs and significance
levels for all three dynamics metrics across the 9 experimental suites. Directional
effects are consistent in all suites for the primary outcome (log-PPL): $C$ is
uniformly positive, $Q$ and $P$ are uniformly negative. Significance levels are
stable across suite types, with minor attenuation in the reasoning suite (likely due
to the lower attractor count $k^*=2$ reducing the dynamic range of $Q$). This
cross-suite consistency supports the generalizability of the regression findings
beyond the anchor model and prompt.

\begin{table}[htbp]
\centering
\caption{Direction of partial regression coefficients for log-PPL across 9
experimental suites. $+$ ($-$) indicates positive (negative) significant coefficient
($p < 0.05$, BH-corrected); $\circ$ indicates non-significant.}
\label{tab:cross_suite}
\small
\begin{tabular}{lccc}
\toprule
\textbf{Suite} & $C$ & $Q$ & $P$ \\
\midrule
DeepSeek-R1 / Factual        & $+$ & $-$ & $-$ \\
DeepSeek-R1 / Reasoning      & $+$ & $-$ & $\circ$ \\
DeepSeek-R1 / Creative       & $+$ & $-$ & $-$ \\
DeepSeek-R1 / Continuation   & $+$ & $-$ & $-$ \\
Gemma-1.1-7B-It              & $+$ & $-$ & $-$ \\
Llama-3.1-8B-Instruct        & $+$ & $-$ & $-$ \\
Mistral-7B-Instruct-v0.3     & $+$ & $-$ & $-$ \\
Qwen2.5-7B                   & $+$ & $-$ & $-$ \\
Qwen2.5-7B-Instruct          & $+$ & $\circ$ & $-$ \\
\bottomrule
\end{tabular}
\end{table}

\subsection{Sanity Check Coefficient Tables}
\label{app:sanity}

Table~\ref{tab:ablation_b} reports partial regression coefficients for the three control
conditions described in §\ref{sec:ablation}: temporal shuffling of hidden states,
random shuffling of indicator--quality pairs, and replacement of hidden states with
i.i.d.\ Gaussian vectors. Under all three controls, the primary continuity coefficient
for log-PPL is reduced to near-zero and loses statistical significance, confirming that
the observed associations depend on the temporal structure and authentic geometry of the
latent trajectories.

\begin{table}[htbp]
\centering
\caption{Ablation C: Sanity Checks — Coefficient (significance)}
\label{tab:ablation_b}
\resizebox{\textwidth}{!}{%
\begin{tabular}{llccc}
\toprule
Configuration & Target & $\beta_C$ & $\beta_Q$ & $\beta_P$ \\
\midrule
  \textbf{DMET (Full)} & log\_ppl & -0.9589*** & -0.4198** & -0.0050*** \\
   & spelling & -0.0041** & 0.0089* & 0.0003*** \\
   & diversity & -0.0955*** & -0.0670 & -0.0007* \\
   & grammar & 0.0012 & -0.0030 & 0.0002 \\
   & coherence & 0.0646*** & -0.0798*** & 0.0008*** \\
\midrule
  Shuffle temporal order & log\_ppl & -0.5526*** & -0.1768 & -0.0061*** \\
   & spelling & -0.0045*** & 0.0085* & 0.0003*** \\
   & diversity & -0.0060 & -0.0225 & -0.0010** \\
   & grammar & -0.0002 & -0.0036 & 0.0002 \\
   & coherence & 0.0382*** & -0.0940*** & 0.0009*** \\
\midrule
  Gaussian random states & log\_ppl & -9.3779 & 0.3171 & -0.0198*** \\
   & spelling & 0.2882 & 0.0121 & -0.0000 \\
   & diversity & -1.3091 & 0.0631 & -0.0015 \\
   & grammar & 0.2007 & -0.0362 & 0.0002 \\
   & coherence & 1.3222 & -0.1326* & 0.0016** \\
\midrule
  Shuffle indicator-quality pairs & log\_ppl & -0.0162 & -0.1165 & -0.0002 \\
   & spelling & -0.0006 & 0.0049 & 0.0000 \\
   & diversity & -0.0027 & -0.0405 & -0.0000 \\
   & grammar & -0.0010 & -0.0004 & 0.0000 \\
   & coherence & 0.0000 & 0.0191 & 0.0002 \\
\bottomrule
\end{tabular}}
\end{table}

\subsection{Adaptive Decoding: Trajectory and Monitoring Plots}
\label{app:adaptive_trajectories}

\begin{figure}[htbp]
\centering
\begin{subfigure}{0.48\linewidth}
\centering
\includegraphics[width=\linewidth]{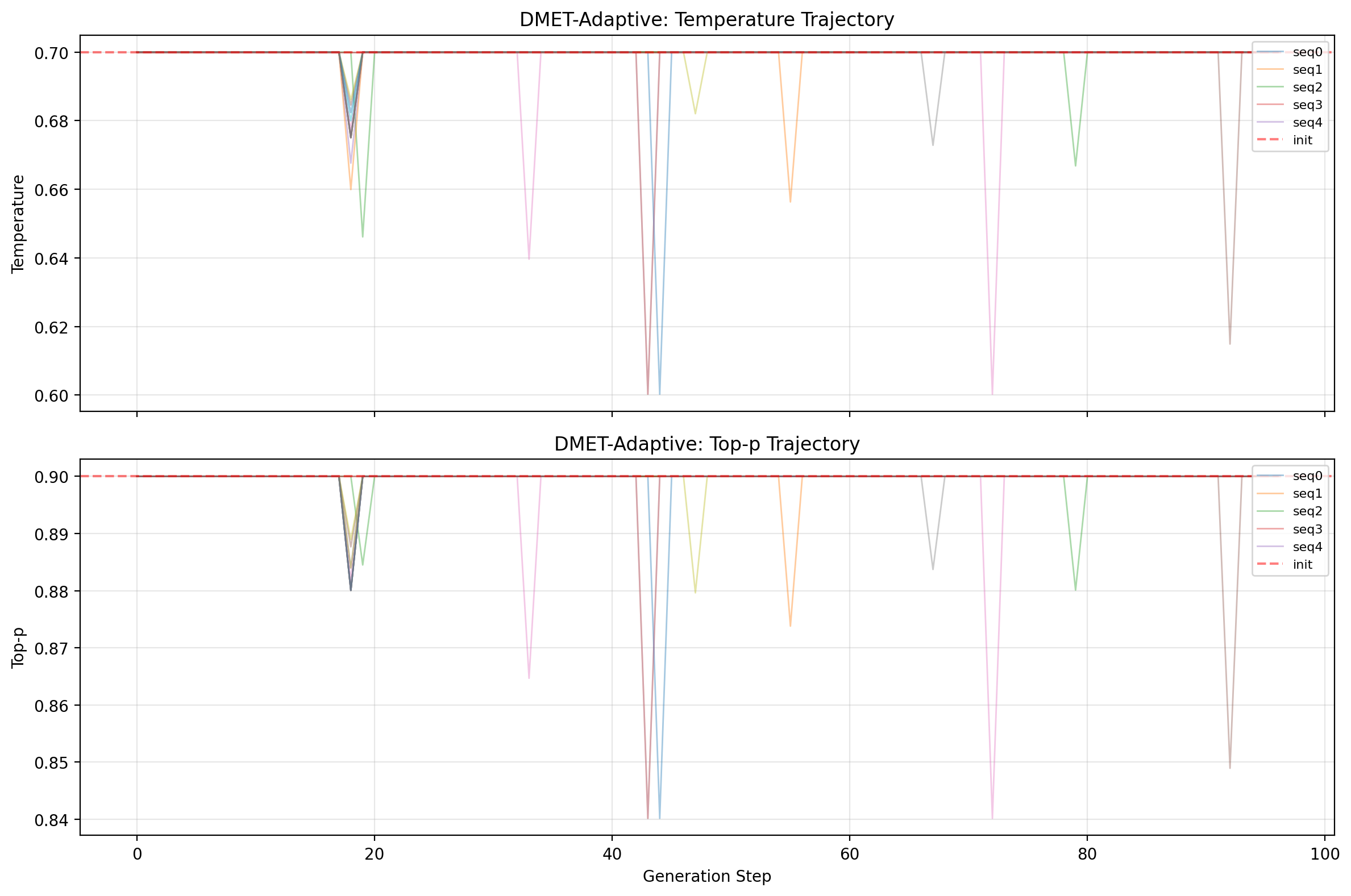}
\caption{Trajectory evolution: adaptive (blue) vs.\ fixed baseline (grey).}
\label{fig:adaptive-traj}
\end{subfigure}
\hfill
\begin{subfigure}{0.48\linewidth}
\centering
\includegraphics[width=\linewidth]{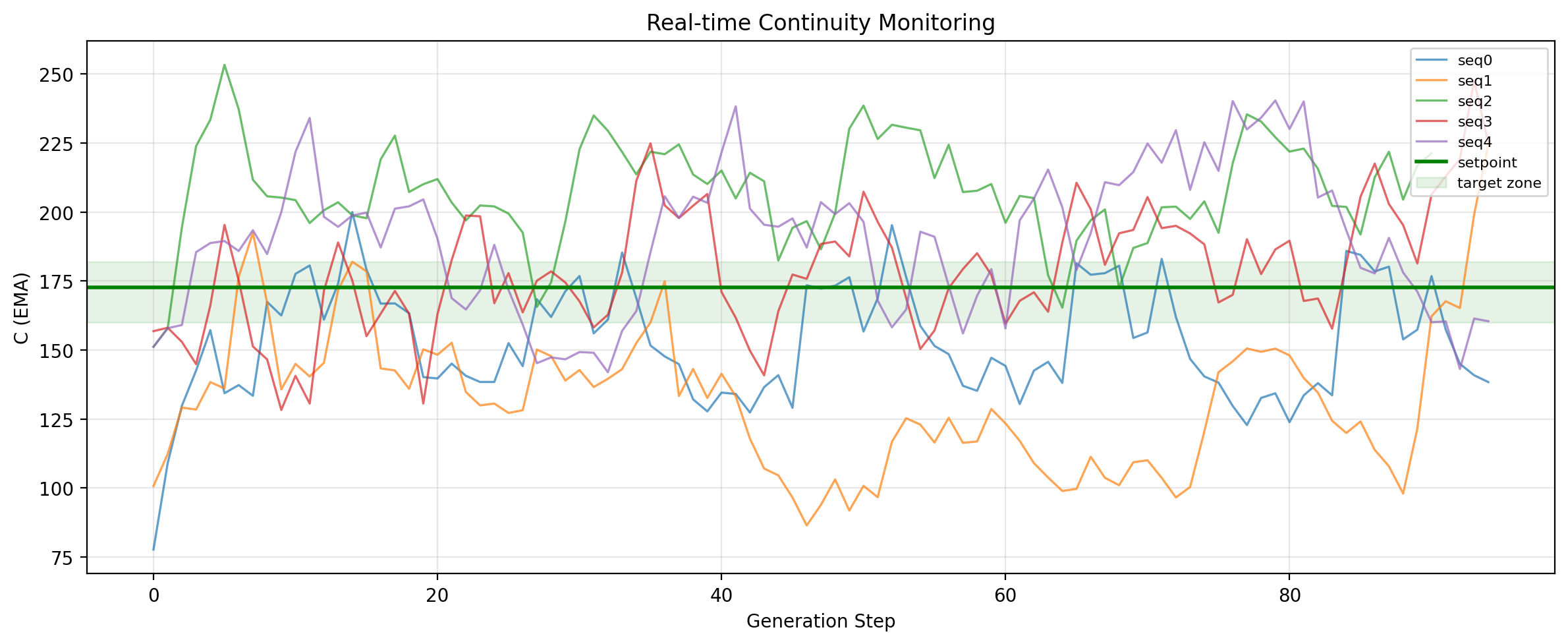}
\caption{Real-time continuity monitor $C_t$ with controller intervention markers.}
\label{fig:adaptive-c}
\end{subfigure}
\caption{Adaptive decoding internals. Left: PCA-space trajectories show that the
adaptive controller keeps generation paths closer to attractor basins. Right:
the continuity monitor triggers $\tau$-reduction at instability events (red markers),
after which $C_t$ visibly decreases, linking controller behavior to trajectory outcomes.}
\label{fig:adaptive-combined}
\end{figure}

Figure~\ref{fig:adaptive-traj} shows the evolution of hidden-state trajectories under
adaptive versus fixed-parameter decoding in the PCA-reduced space. Adaptive trajectories
are visibly more compact and concentrated near attractor basins, consistent with the
controller's objective of minimizing $C$ excursions. Figure~\ref{fig:adaptive-c} shows
the real-time continuity monitor $C_t = \|\mathbf{h}_t - \mathbf{h}_{t-1}\|_2$ during
a representative adaptive decoding run: the controller fires (reducing $\tau$) at
time steps where $C_t$ exceeds the threshold, visibly damping subsequent trajectory
excursions and confirming that quality improvements are causally linked to trajectory
regulation rather than to incidental parameter changes.

\subsection{Trajectory Evolution Analysis}
\label{app:traj_evolution}

\begin{figure}[htbp]
\centering
\includegraphics[width=\linewidth]{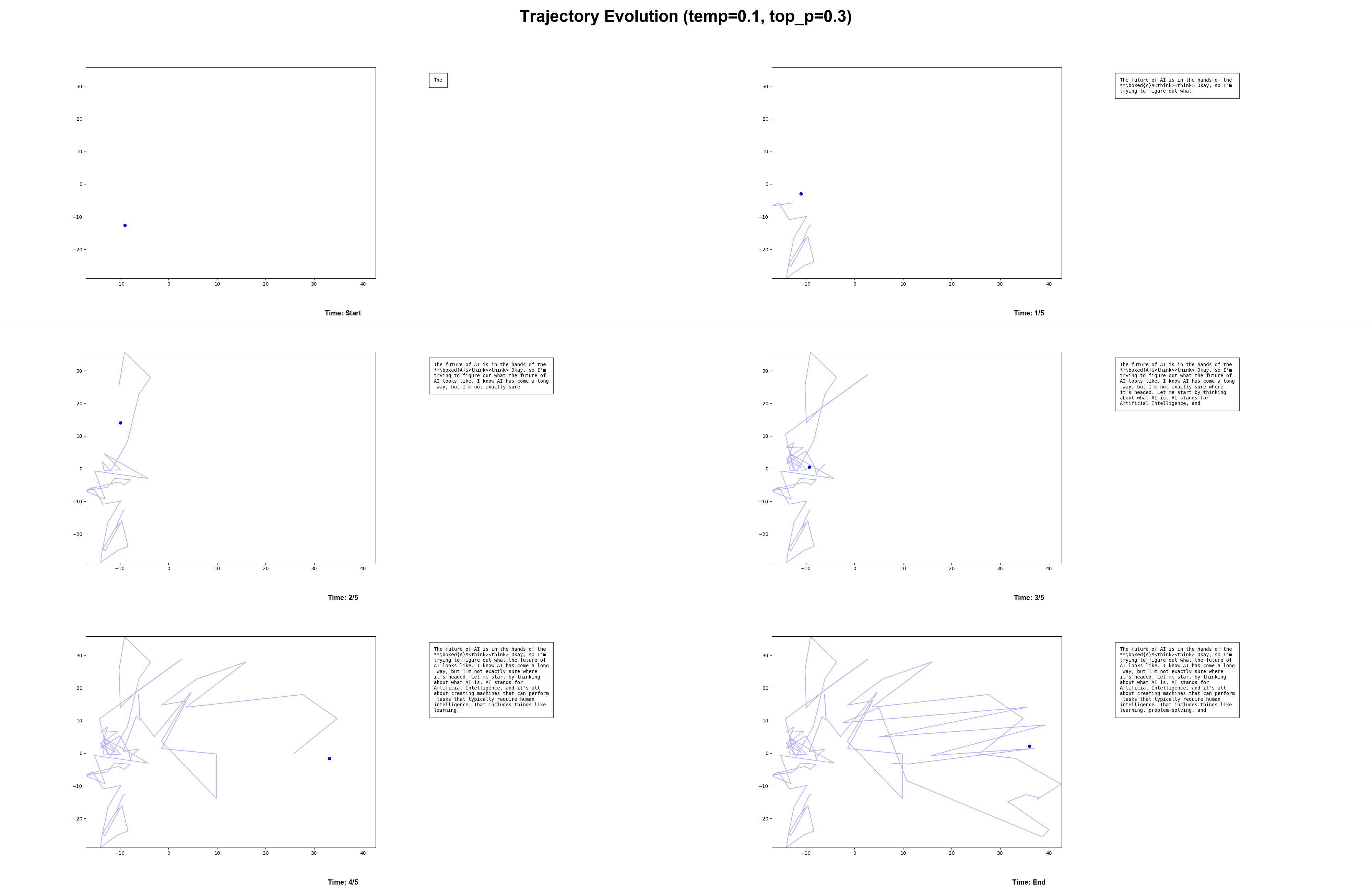}
\caption{Single-sequence trajectory in 2D PCA space
($\tau = 0.1$, top-$p = 0.3$). Low temperature produces a compact, near-linear
path converging rapidly to the dominant attractor.}
\label{fig:traj1}
\end{figure}

\begin{figure}[htbp]
\centering
\includegraphics[width=\linewidth]{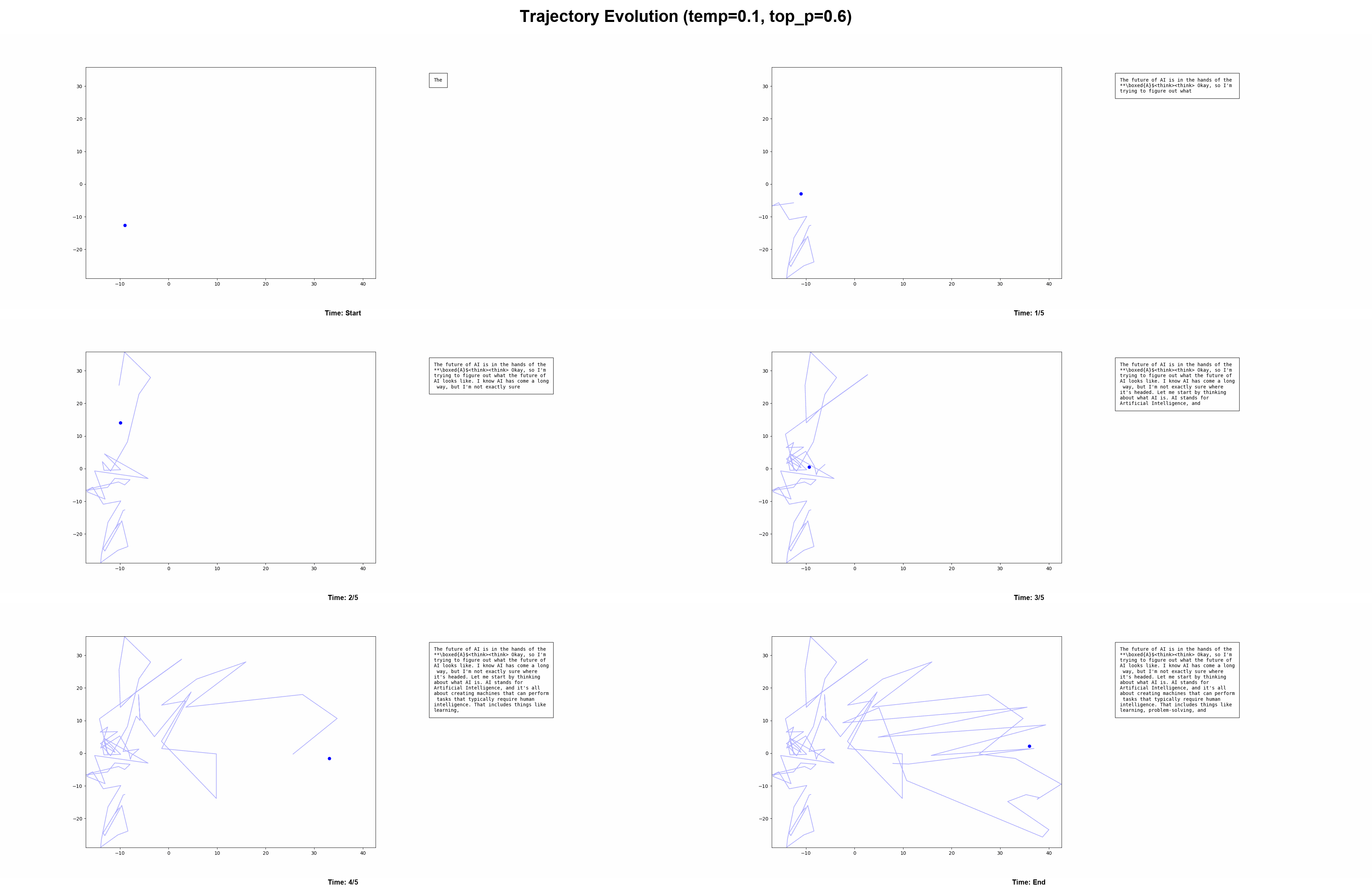}
\caption{Single-sequence trajectory in 2D PCA space
($\tau = 0.1$, top-$p = 0.6$). Slightly wider top-$p$ marginally broadens the
trajectory without disrupting convergence.}
\label{fig:traj2}
\end{figure}

\begin{figure}[htbp]
\centering
\includegraphics[width=\linewidth]{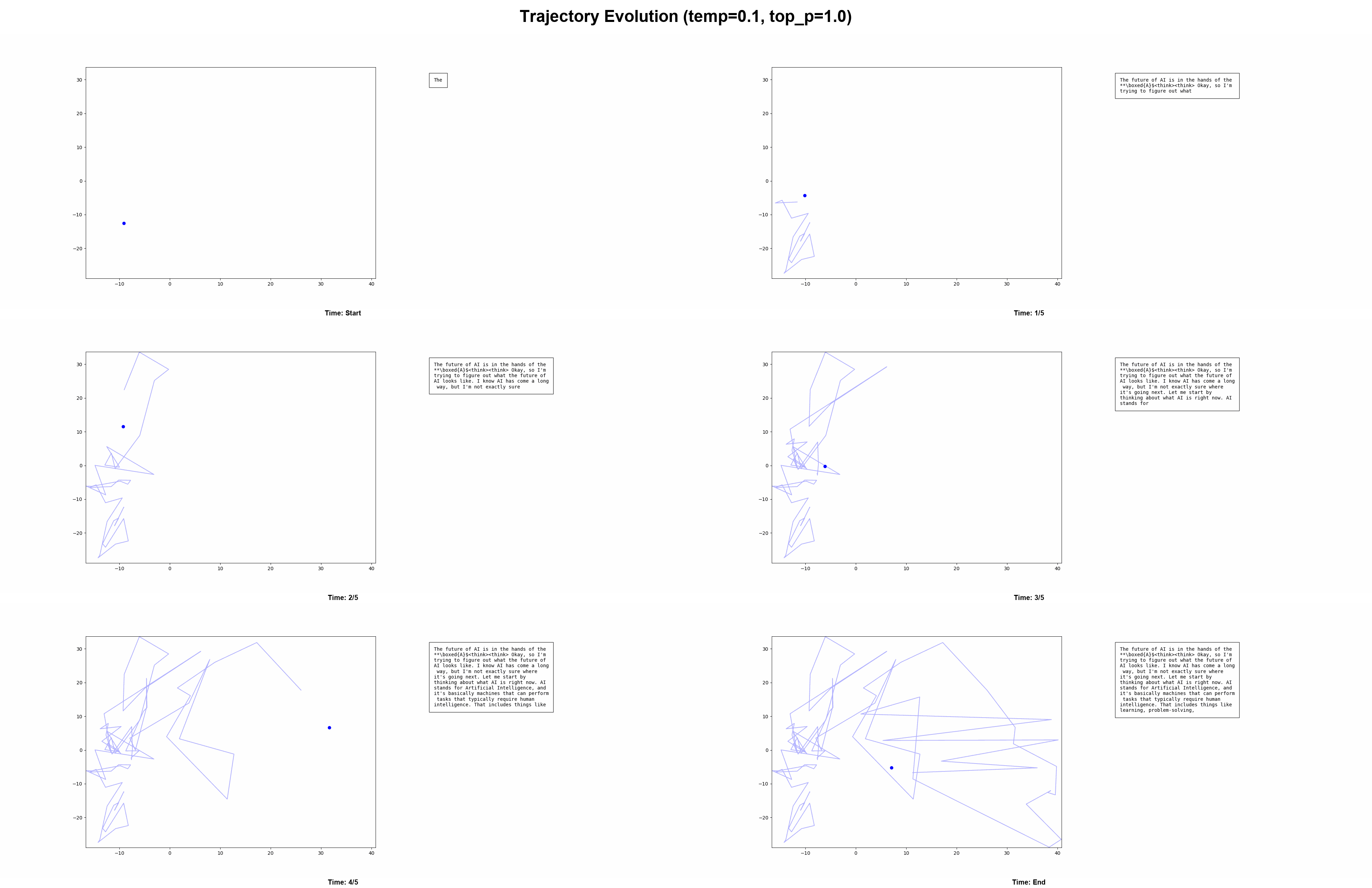}
\caption{Single-sequence trajectory in 2D PCA space
($\tau = 0.1$, top-$p = 1.0$). Full nucleus sampling with low temperature
preserves fluency while permitting limited topological exploration.}
\label{fig:traj3}
\end{figure}

\begin{figure}[htbp]
\centering
\includegraphics[width=\linewidth]{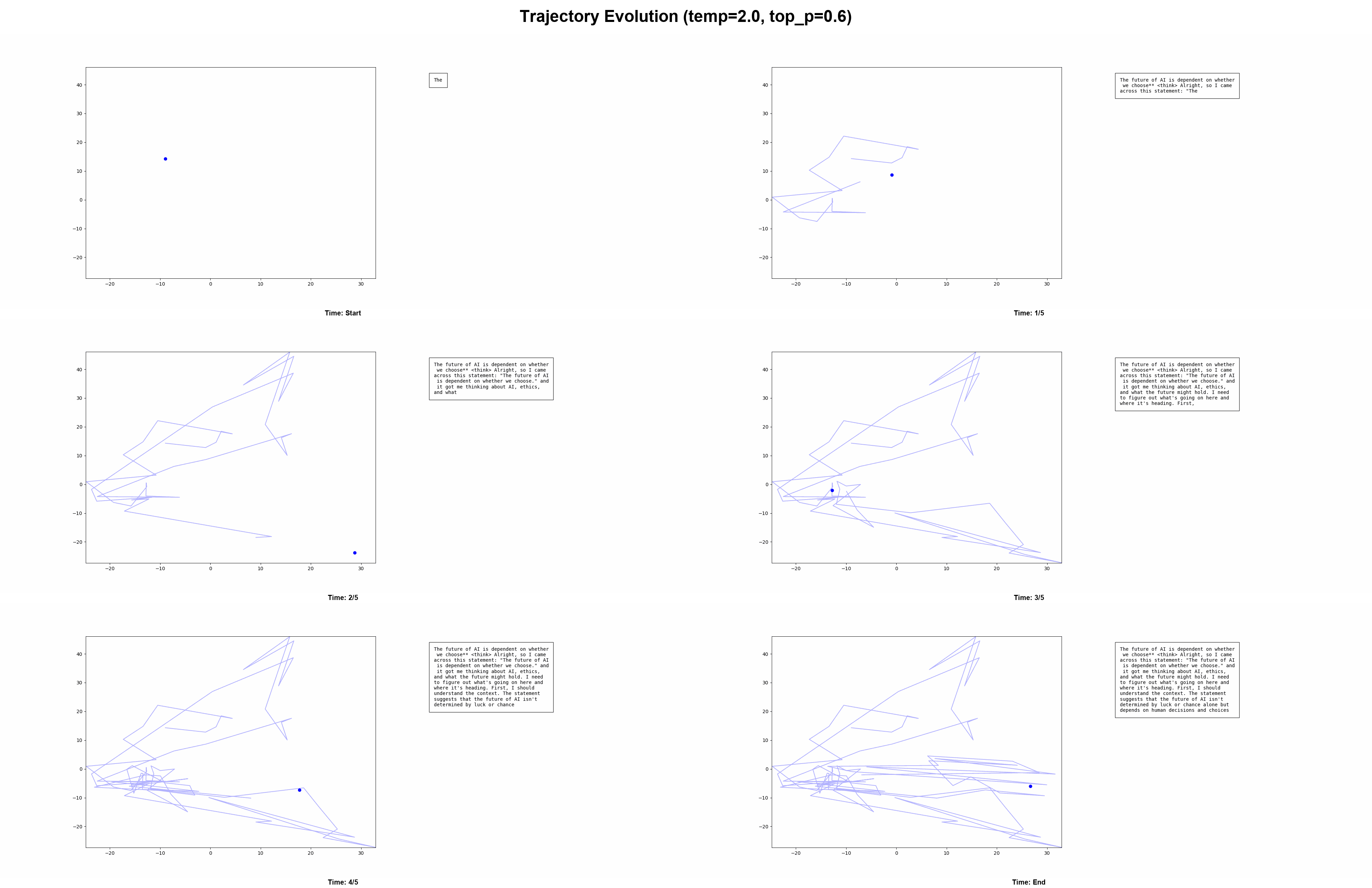}
\caption{Single-sequence trajectory in 2D PCA space
($\tau = 2.0$, top-$p = 0.6$). High temperature produces wide, irregular
excursions with weaker attractor convergence, consistent with the stochastic
perturbation interpretation of Eq.~\eqref{eq:sde_approx}.}
\label{fig:traj4}
\end{figure}

Figures~\ref{fig:traj1}--\ref{fig:traj4} depict single-sequence trajectory evolution
in 2D latent space under four representative decoding configurations. A consistent
three-phase pattern emerges: an \emph{initial phase} (tokens 1--10) in which the
trajectory explores a local neighborhood, reflecting uncertainty about semantic
direction; an \emph{expansion phase} (tokens 30--60) in which the trajectory enters
new regions corresponding to topic development; and a \emph{convergence phase}
(tokens 70--100) in which the trajectory stabilizes near an attractor basin,
corresponding to natural content closure. This phase structure is qualitatively
consistent across temperature and top-$p$ settings, though the expansion phase is
more pronounced at higher temperatures.

\end{document}